# Position Detection and Direction Prediction for Arbitrary-Oriented Ships via Multiscale Rotation Region Convolutional Neural Network

Xue Yang, *Student Member, IEEE*, Hao Sun, Xian Sun, Menglong Yan, Zhi Guo, and Kun Fu

*Abstract*—Ship detection is of great importance and full of challenges in the field of remote sensing. The complexity of application scenarios, the redundancy of detection region, and the difficulty of dense ship detection are all the main obstacles that limit the successful operation of traditional methods in ship detection. In this paper, we propose a brand new detection model based on multiscale rotational region convolutional neural network to solve the problems above. This model is mainly consist of five consecutive parts: Dense Feature Pyramid Network (DFPN), adaptive region of interest (ROI) Align, rotational bounding box regression, prow direction prediction and rotational nonmaximum suppression (R-NMS). First of all, the low-level location information and high-level semantic information are fully utilized through multiscale feature networks. Then, we design adaptive ROI Align to obtain high quality proposals which remain complete spatial and semantic information. Unlike most previous approaches, the prediction obtained by our method is the minimum bounding rectangle of the object with less redundant regions. Therefore, rotational region detection framework is more suitable to detect the dense object than traditional detection model. Additionally, we can find the berthing and sailing direction of ship through prediction. A detailed evaluation based on SRSS and DOTA dataset for rotation detection shows that our detection method has a competitive performance.

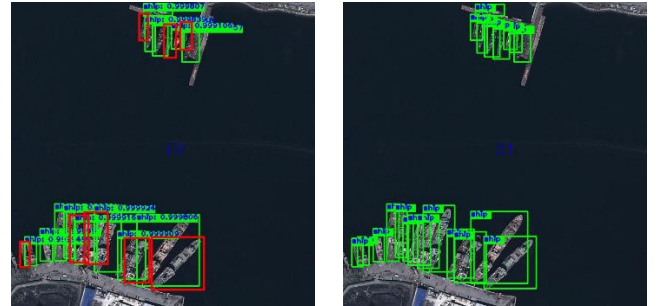

(a) Horizontal region detection: detection result (first column) and ground-truth (second column)

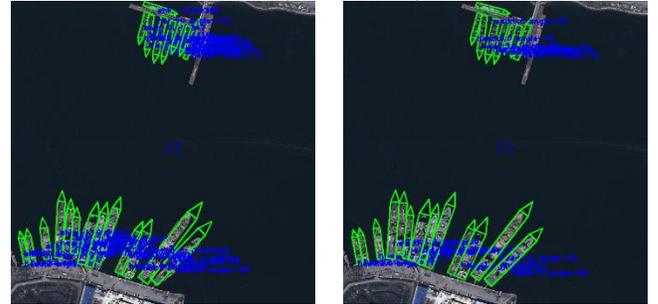

(b) Rotational region detection: detection result (first column) and ground-truth (second column)

Fig. 1. Rotational region detection algorithm perfectly solves the problem of traditional detection algorithm in dense object detection. The green, red bounding boxes represent predictions boxes and missing prediction boxes respectively. (a) Horizontal region detection. (b) Rotational region detection.

## I. INTRODUCTION

WITH the development of remote sensing technology, high-resolution remote sensing images can be easily obtained. Automatic ship detection has been playing a significant role in the field of remote sensing for a long time and has made a great progress in promoting national defense construction, port management, cargo transportation, and maritime rescue [1]. Simultaneously, the information of ship's berthing and sailing direction are also of huge significance. However, the characteristics of the large aspect ratio make ship detection become more difficult than other object detections, such as vehicles [2]-[6], buildings [7]-[12] and aircrafts [13]-[18]. What's more, the complexity of application scenarios, the redundancy of detection region, and the difficulty of dense ship detection have posed a great challenge for ship detection and direction prediction.

In recent years, deep learning has achieved great success in computer vision [19]-[24]. Much attention has been paid to object detection based on deep learning methods and has made great achievement. Region proposals with convolutional neural network (RCNN) [25] provide an excellent pipeline for object detection. Although RCNN has some obvious defects in computation speed and storage space, its detection results are far better than traditional detection methods. Fast-RCNN [26] significantly improve the efficiency of detection and effectively reduce the storage by using shared computing. Faster-RCNN

This work was supported in part by the National Natural Science Foundation of China under Grants 41501485, Grants 41701508 and Grants 61725105. *(Correspondence author: Kun Fu.)*

X. Yang, H. Sun, X. Sun, M. L. Yan, Z. Guo, and K. Fu are with Key Laboratory of Technology in Geo-spatial Information Processing and Application System, Institute of Electronics, Chinese Academy of Sciences, Beijing 100190, China.

X. Yang, and K. Fu are with School of Electronic, Electrical and Communication Engineering, University of Chinese Academy of Sciences, Beijing 100049, China (e-mail: fukun@mail.ie.ac.cn).

[27] adopts a trainable region proposal network (RPN) instead of Selective Search method to achieve end-to-end training while improving detection efficiency and accuracy. It is consist of two stage: region proposal and region classification.

The methods above are known as horizontal region detection which are suitable for natural scene detection but not for satellite remote sensing ship detection. In satellite remote sensing images, the ships have a large aspect ratio and are often densely arranged in complex scenes. Once the ship is inclined, the redundant regions of the horizontal bounding box and the regions of overlap between the ships will be relatively large. The disadvantage of this situation is obvious and disastrous. Specifically, complex scenes often contain many noise objects, which greatly affect the performance of the ship detection. In addition, large redundant regions introduce a lot of noise, causing the feature information to be interfered or even submerged. As shown in Fig. 1(a), a large object overlap region causes the object to be discarded after the operation of nonmaximum suppression (NMS). To address these problems above, we propose a new, end-to-end, rotational-region-based object detection framework for ship detection in high-resolution satellite images which can handle different complex scenarios, detect intensive objects, and reduce redundant detection regions, as illustrated in Fig. 1(b). Moreover, our framework can predict the berthing and sailing direction of ship, which cannot be achieved by the horizontal region detection method.

Our framework mainly is consist of five consecutive parts: Dense Feature Pyramid Network (DFPN), adaptive region of interest (ROI) Align, rotational bounding box regression, prow direction prediction and rotational nonmaximum suppression (R-NMS). Compared with detection methods based on convolutional neural network (CNN), our framework is more suitable for ship detection and has achieved more promoting performance.

The main contributions of this paper are as follows:

1) *DFPN*

   We design a new multiscale feature fusion network called DFPN, which can effectively integrate the low-level location information and high-level semantic information to provide more advanced features for object detection.

2) *Adaptive ROI Align*

   Adaptive ROI Align is proposed in this paper to mitigate the effects of redundant noise regions in the proposals and keep the completeness of semantic and spatial information.

3) *Prow Direction Prediction*

   The berthing and sailing direction of ship can be found through prediction. This method is simple but effective, with a high prediction accuracy.

4) *R-NMS*

   In order to obtain more accuracy prediction results, we propose R-NMS which has more stringent constraints.

The rest of this paper is organized as follows. Section II we briefly review related work on object detection. Section III

introduces the details of the proposed method. Section IV presents experiments conducted on a remote sensing dataset to validate the effectiveness of the proposed framework. Finally, section V discusses and concludes the results of the proposed method.

## II. RELATED WORK

Ship detection has been investigated by a wide variety of methods in recent years. In this section, we briefly review the existing machine-learning-based ship detection algorithm and deep-learning-based ship detection algorithm.

In the past few years, some machine-learning-based methods have been proposed for ship detection [28]–[32]. Yu, Y.D. *et al.* and Zhu, C. *et al.* [33]-[34] propose features of texture and shape by sea-land segmentation, then an algorithm such as the contrast box algorithm or semi-supervised hierarchical classification is used to get the candidate object region. Bi F *et al.* [35] use a bottom-up visual attention mechanism to select prominent candidate regions throughout the detection scene. Yang *et al.* [36] propose a novel detection framework via sea surface analysis to solve the task of automatic ship detection in high-resolution optical satellite images with various sea surfaces. This proposed method first use two new features to analyze whether the sea surface is homogeneous or not. Then, they propose a linear function combining pixel and region characteristics to select ship candidates. Finally, false alarms are filtered by adopting compactness and length-width ratio. Shi *et al.* [37] present a method to detect ships in a "coarse-to-fine" manner. Specially, they convert an optical image into a hyperspectral form by adopting anomaly detector and local shape features, then extract ships through hyperspectral algorithm. Corbane *et al.* [38] present a complete processing chain for ship detection based on statistical methods, mathematical morphology and other signal-processing techniques such as the wavelet analysis and Radon transform. This automatic ship detection model is used to complement existing regulations, especially the fishing control system.

Although these machine-learning-based ship detection algorithm above have shown promising performance, they have poor practicability in complex scenarios. With the application of deep CNN in object detection [39]-[43], deep-learning-based ship detection algorithm are also widely used in remote sensing ship detection. Kang M *et al.* [44] take the objects proposals generated by Faster R-CNN for the guard windows of CFAR algorithm, then pick up the small objects, thus reevaluating the bounding boxes which have relative low classification scores in detection network. Zhang R *et al.* [45] propose a new ship detection model based on CNN which is called SCNN, fed with specifically designed proposals extracted from the ship model combined with an improved saliency detection method. Kang M *et al.* [46] build a contextual region-based CNN with multilayer fusion for SAR ship detection, which is an elaborately designed deep hierarchical network, and composed of a RPN with high network resolution and an object detection network with contextual features. Tang *et al.* [47] adopt compressed domain for fast ship candidate extraction, while DNN is exploited for high-level feature representation and classification, and ELM is used for efficient feature pooling and decision making.

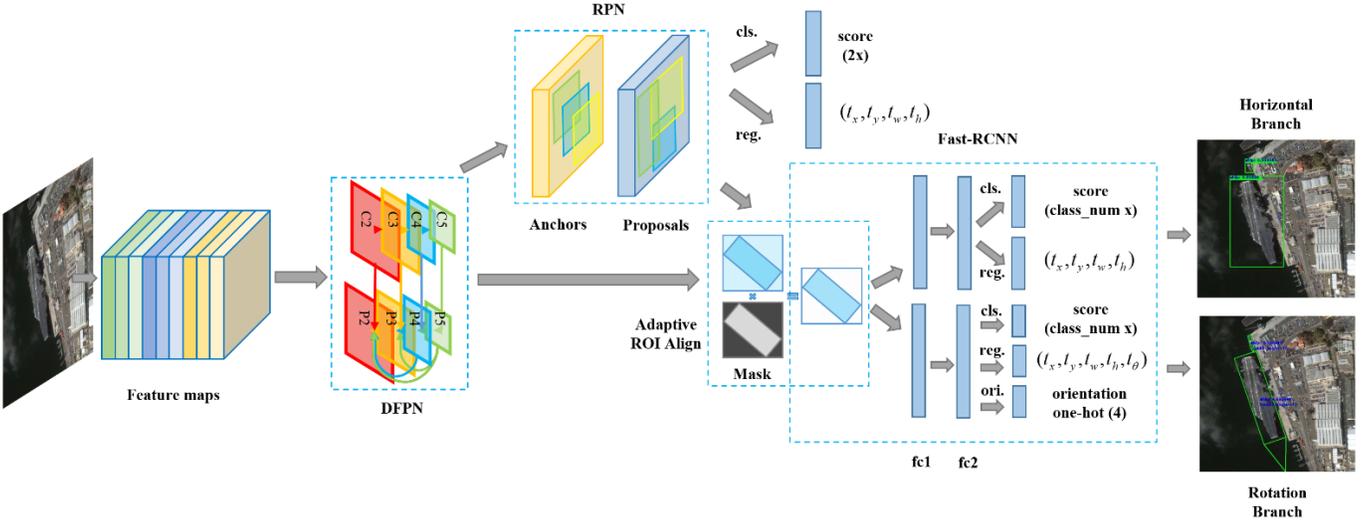

Fig. 2. Overall framework of rotational region ship detection. This framework mainly consists of five consecutive parts: DFPN, adaptive ROI Align, rotational bounding box regression, prow direction prediction and R-NMS.

These methods above are mostly based on horizontal region detection. Detection results tend to have very large redundant regions, and is not conducive to NMS operation. We propose a novel object detection model based on multiscale rotation region CNN which effectively integrates the low-level location information and high-level semantic information. Meanwhile, this method mitigate the effects of redundant noise regions in the proposals and get rotational bounding box with prow direction. Compared with other deep-learning-based ship detection framework, our method can achieve state-of-the-art detection performance, even in dense scenes.

## III. PROPOSED METHOD

The overall framework of our rotational region ship detection method is illustrated in Fig. 2. DFPN, adaptive ROI Align, rotational bounding box regression, prow direction prediction and R-NMS are the five important components of our method. Firstly, DFPN [1] is an effective multiscale feature fusion network which enhances feature propagation, encourages feature reuse, and ensures the effectiveness of detecting multiscale objects. Then, we get proposals from the RPN to provide high-quality region proposals for the next stage. In order to keep the completeness of semantic and spatial information, we design an adaptive ROI Align to mitigate the effects of redundant noise regions in the proposals. Furthermore, compared with traditional framework, second stage of our model has horizontal and rotational branches that respectively predict the horizontal bounding box and the rotational bounding box. Meanwhile, the rotation branch can also predict the berthing and sailing direction of ship. Finally, we use R-NMS which has more stringent constraints so as to obtain the final prediction.

### A. Dense Feature Pyramid Network

The low-level location information and high-level semantic information are very important to object detection. The feature pyramid [32] is an effective multiscale method to fuse multilevel information, so we adopt a multiscale feature pyramid connection which we called DFPN. Fig. 3 depicts the structure of this densely connected multiscale pyramid feature fusion network.

In this paper, we use ResNet [48] as backbone and choose the last layer of each residual block as the feature maps $\{C_2, C_3, C_4, C_5\}$ in the bottom-up feedforward network. According to the residual network structure, the strides of each feature map correspond to $\{4,8,16,32\}$ pixels. In the top-down network, we get higher resolution features $\{P_2, P_3, P_4, P_5\}$ through lateral connections and dense connections. We set the number of channels for all feature maps to 256 so as to reduce the number of parameters. The specific definition is as follows:

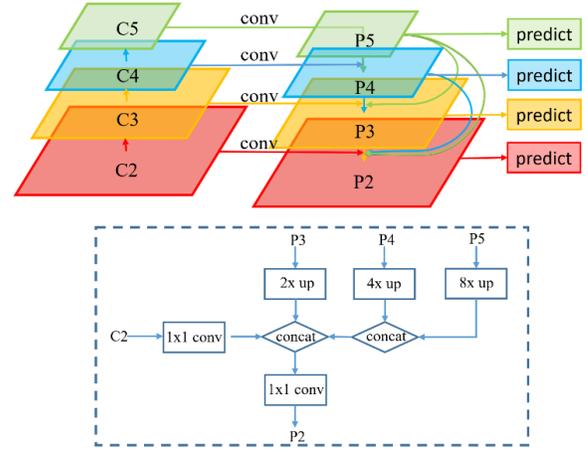

Fig. 3. A multiscale feature pyramid connection. Each feature map is densely connected, and merged by concatenation

$$P_5 = Conv_{1\times1}(C_5)$$
$$P_i = Conv_{3\times3}[\sum_{j=i+1}^{5} Upsample(P_j) \oplus Conv_{1\times1}(C_i)] \quad (1)$$

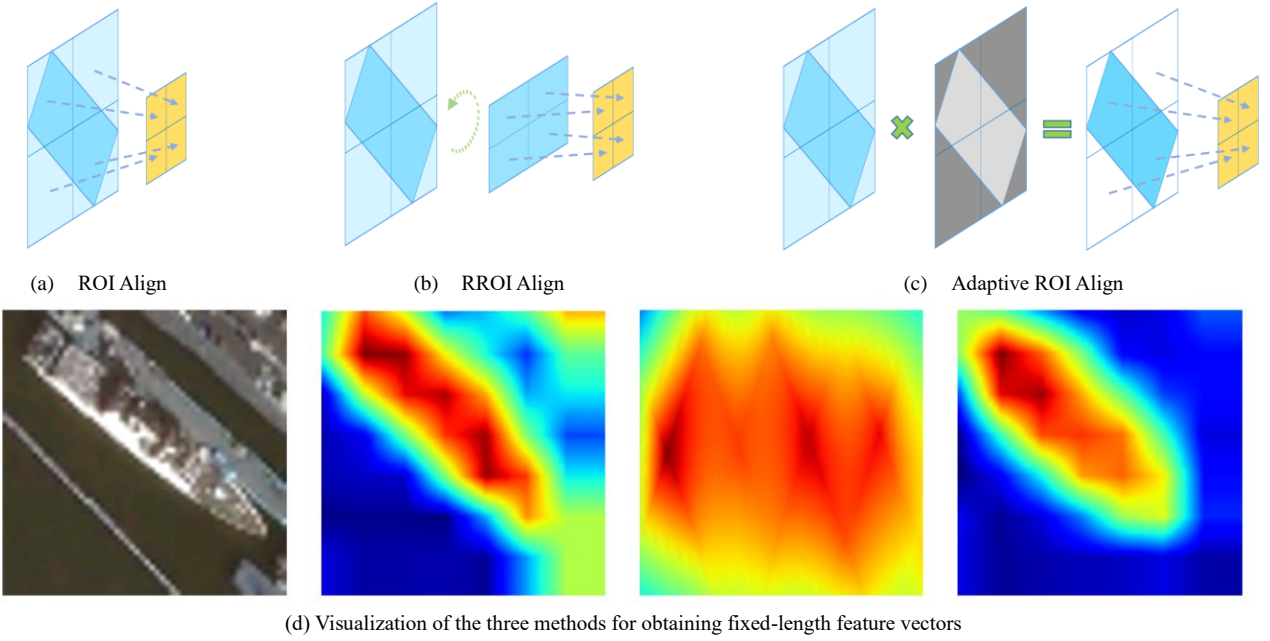

(a) ROI Align      (b) RROI Align      (c) Adaptive ROI Align

(d) Visualization of the three methods for obtaining fixed-length feature vectors

Fig. 4. Structure and visualization of three Align methods. (a) ROI Align. (b) RROI Align. (c) Adaptive ROI Align. (d) Visualization of the three methods for obtaining fixed-length feature vectors.

where $P_i$ is the fused feature map corresponding to $C_i$. $Conv_{k \times k}(.)$ represents the convolution operation, and k is the size of the convolution kernel. $Upsample(.)$ represents nearest neighbor up-sampling in this paper. $\oplus$ represents the operation of concatenation.

We assign five scales $\{32,64,128,256,512\}$ pixels to $\{P_2, P_3, P_4, P_5, P_6\}$ respectively ($P_6$ is simply a stride two subsampling of $P_5$). Taking into account of the characteristics of ships, the ratios of anchor are $\{1:7, 1:5, 1:3, 1:2, 1, 2, 3, 5, 7\}$. Each feature point for each feature map will generate 9 anchors $(1 \times 9)$, 45 outputs $(5 \times 9)$ for each regression layer, and 18 outputs $(2 \times 9)$ for each classification layer. A large number of experiments show that DFPN has a great feature of fusion, and significantly improve the detection performance

### B. Adaptive ROI Align

The large aspect ratio is a major feature of the ship. However, once the ship is inclined, the redundant regions of the proposal are relatively large. A lot of noise will reduce the quality of feature extraction, or even submerged features. Fig. 4(a-c) shows three methods for obtaining fixed-length feature vectors: ROI Align [1], RROI Align [49], and adaptive ROI Align. It is obvious that ROI Align is accompanied by a lot of noise, causing the target features to be overwhelmed. Although RROI removes all the noise through affine transformation, it loses the spatial information of the object. We designed Adaptive ROI Align, a method that automatically filters noise regions by introducing a mask. This mask is trainable and is obtained by convolving the proposals. Adaptive ROI Align makes the spatial information be retained, while leaving a small amount

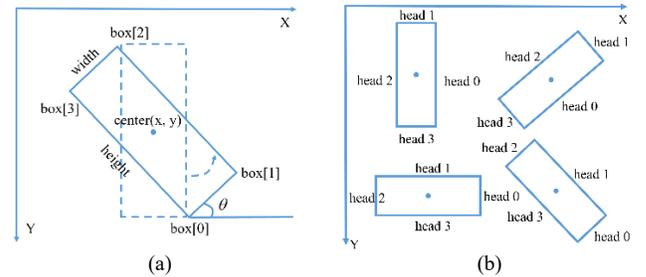

(a)              (b)

Fig. 5. Representation of rotational bounding box and prow direction. (a) Representation of rotational bounding box. (b) Prow direction.

of noise to improve the stability of the network.

Fig. 4(d) visualizes the three methods for obtaining fixed-length feature vectors. As we can see, coast is the main noise in the image, ROI Align can't completely remove it. RROI Align lost spatial information and produced the feature deformation at the same time, which are not conducive to angle regression and prow direction prediction in the second stage. The adaptive ROI Align better solves the problem of the method above, and have access to high-quality feature maps.

### C. Prow Direction Prediction

The horizontal bounding box is represented by the upper left corner and the lower right corner, such as $(x_{min}, y_{min}, x_{max}, y_{max})$. However, this representation lacks direction information, thus we use five variables $(x, y, w, h, \theta)$ to redefine the arbitrary rotated bounding box. As shown in Fig. 5(a), rotation angle $\theta$ is the angle at which the horizontal axis (x-axis) rotates counterclockwise to the first edge of the encountered rectangle. At the same time, we define this side as the width $w$, the other is the height $h$. It is worth mentioning that the range of angles

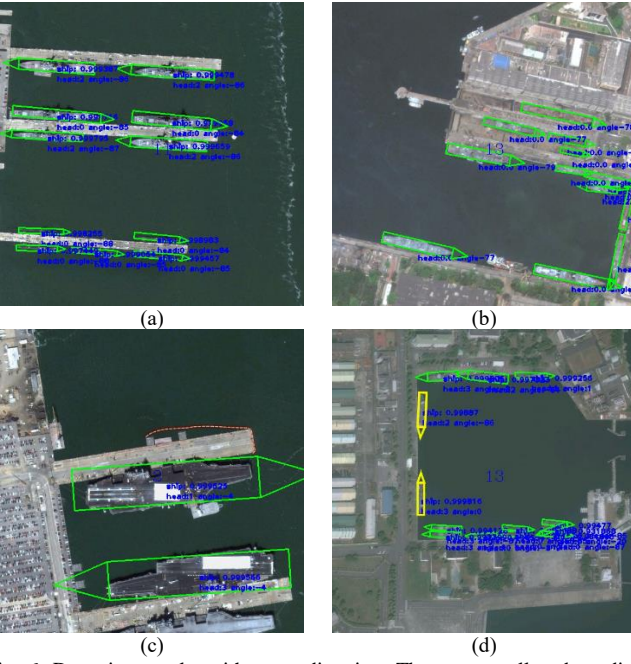

(a)              (b)

(c)              (d)

Fig. 6. Detection results with prow direction. The green, yellow bounding boxes represent predictions boxes and error prow direction prediction boxes respectively. (a) Scene one: The ships are arranged side by side but in different directions. (b) Scene two: The ships berth in the harbor and dock. (c) Scene three: Large objects. (d) Scene four: Small objects.

is $[-90, 0)$.

From the definition of rotating bounding box we can see that rotation angle $\theta$ cannot represent the prow direction of the object, but the prow direction is certainly in the direction of the four sides of the minimum bounding rectangle. In the light of these facts, we label the four sides of the rotation bounding box counterclockwise, as shown in Fig. 5(b). Meanwhile, we predict the berthing and sailing direction of the ship in the rotation branch.

Fig. 6 shows the detection results of four different scenarios, we have achieved amazing performance with such simple predictions. Part of the wrong prediction is to judge the stern of the prow. In our opinion, this network first learned that the prow must be in the long edge direction and then judged which side of the long edge is the prow.

### D. Loss Function

During training of the RPN, each anchor is assigned a binary class label and five parametric coordinates. The feature maps were input to the RPN network through a $3 \times 3$ convolutional layer, followed by two sibling $1 \times 1$ convolution layers for regression and classification. We need to find positive and negative samples from all anchors, which we call a mini-batch. The positive sample anchors need to satisfy the following conditions: the Intersection-over-Union (IoU) overlap between an anchor and the ground-truth is greater than 0.6. The negative samples are defined as: IoU overlap less than 0.25. The total number of positive and negative samples is 256, the ratio is 0.5. Similar to the RPN stage, the second stage classifies each proposal and assign five parametric coordinates to regress the final rotational bounding box. The ratio of positive and negative samples in mini-batch is 0.5, the threshold is 0.5 and the total number of samples is 128.

After adding the angle information, rotational bounding box can locate the object more accurately. The regression of rotational bounding box is defined as follows:

$$
\begin{aligned}
&t_x = (x - x_a) / w_a, t_y = (y - y_a) / h_a, \\
&t_w = \log(w / w_a), t_h = \log(h / h_a), \\
&t_\theta = \theta - \theta_a + k\pi / 2
\end{aligned} \tag{2}
$$

$$
\begin{aligned}
&t^*_x = (x^* - x_a) / w_a, t^*_y = (y^* - y_a) / h_a, \\
&t^*_w = \log(w^* / w_a), t_h = \log(h^* / h_a), \\
&t^*_\theta = \theta^* - \theta_a + k\pi / 2
\end{aligned} \tag{3}
$$

where $x, y, w$, and $h$ denote the box's center coordinates and its width and height. Variables $x$, $x_a$, and $x^*$ are for the predicted box, anchor box, and ground-truth box, respectively (likewise for $y, w, h$). The parameter $k \in Z$ keeps $\theta$ in the range $[-90, 0)$. In order to keep the bounding box in the same position, $w$ and $h$ need to be swapped when $k$ is an odd number.

We use multitask loss to minimize the objective function, which is defined as follows:

$$
\begin{aligned}
&L(p_i, l_i, u_i^*, u_i, v_i^*, v_i, h_k^*, h_k) \\
&\quad = \frac{1}{N_{cls}} \sum_i L_{cls}(p_i, l_i) + \\
&\quad \lambda_1 \frac{1}{N_{reg-h}} \sum_j p_j L_{reg-h}(u_j^*, u_j) + \\
&\quad \lambda_2 \frac{1}{N_{reg-r}} \sum_k p_k L_{reg-r}(v_k^*, v_k) + \\
&\quad \lambda_3 \frac{1}{N_{reg-r}} \sum_k p_k L_{reg-r}(h_k^*, h_k)
\end{aligned} \tag{4}
$$

where $l_i$ represents the label of the object, $p_i$ is the probability distribution of various classes calculated by the soft-max function, $u_i, v_i$ represent the predicted parameterized coordinate vectors, $u_i^*, v_i^*$ represent the offset of ground-truth. $h_k^*, h_k$ represent the prow direction of ground-truth and prediction respectively. The hyper-parameter $\lambda_1, \lambda_2, \lambda_3$ in (4) control the balance between the four task losses; all experiments use $\lambda_1 = \lambda_2 = 1, \lambda_3 = 10$ in this paper. In addition, the functions $L_{cls}$ and $L_{reg}$ are defined as:

$$
L_{cls}(p, l) = -\log pl \tag{5}
$$

$$
L_{reg}(t_i^*, t_i) = smooth_{L_1}(t_i^* - t_i) \tag{6}
$$

$$
smooth_{L_1}(x) = \begin{cases} 0.5x^2, \text{if } |x| < 1 \\ |x| - 0.5, \text{otherwise} \end{cases} \tag{7}
$$

*E. Rotational Nonmaximum Suppression*

NMS is to obtain high quality bounding boxes with small IoU overlap. When ships are densely arranged, the traditional NMS often faces such a dilemma that the bounding box has a large IoU overlap. Therefore, IoU computation on axis-aligned bounding box may lead to an inaccurate IoU of skew interactive bounding box and further ruin the bounding box prediction. An implementation for Skew IoU computation [49] with thought to triangulation is proposed to deal with this problem.

The sensitive relationship between IoU overlap and rotation angle often affect the detection results. For example, for a ship with aspect ratio of 1: 7, the IoU is only 0.38 when the angles differ by 15 degrees. Therefore, we design R-NMS which has two constraints: (a) preserve the prediction results with IoU less than 0.7; (b) if the IoU is in the range of $[0.3, 0.7]$, discard the prediction results that the angle difference is greater than $15°$.

## IV. EXPERIMENTS

In this section, we will introduce our dataset first. Then we present several groups of comparative experiments to explore the detection performance of the proposed framework. All experiments are conducted on a computer with an NVIDIA GeForce GTX 1080 GPU, and 8 GB of memory.

*A. Dataset and Settings*

In the experiments, we evaluate the proposed framework on two data sets designed for remote sensing image rotation region detection, one of which is a satellite remote sensing ship image dataset (SRSS) that we have collected and labeled, and the other is a publicly available data set named DOTA [50].

SRSS is collected publicly from Google Earth with 50 large scene images sized $10,000 \times 10,000$ pixels, covering 25 square kilometers. The resolution of satellite remote sensing images is 0.5 meters. In addition, these satellite remote sensing images have the tri-band information (include red, green, and blue) after geometric correction. Geotif is the format of satellite image with latitude and longitude information. The images contain scenes of civilian ports, naval base, offshore areas, and far seas. The annotation content is a set of contour points starting from the prow. We divide the images into $1,000 \times 1,000$ subimages with an overlap of 0.4, then filter out images that do not contain ships, resulting in about 8000 final images. Meanwhile, the ratio of training set to test is 1:4.

DOTA is a large-scale dataset for object detection in aerial images. It can be used to develop and evaluate object detectors in aerial images. It contains 2806 aerial images from different sensors and platforms. Each image is of the size in the range from about $800 \times 800$ to $4000 \times 4000$ pixels and contains objects exhibiting a wide variety of scales, orientations, and shapes. These DOTA images are then annotated by experts in aerial image interpretation using 15 common object categories. The fully annotated DOTA images contains 188, 282 instances, each of which is labeled by an arbitrary quadrilateral. In order to ensure that the training data and test data distributions approximately match, half of the original images were randomly selected as the training set, 1/6 as validation set, and 1/3 as the testing set. We also divide the images into subimages as the SRSS data set does.

All experiments were implemented on the deep learning framework, tensorlfow [51]. We use the pretraining model ResNet-101 to initialize the network. For SRSS dataset, we train a total of 40 k iterations, with a learning rate of 0.001 for the first 20 k iterations, 0.0001 for the next 10 k iterations, and 0.00001 for the remaining 10 k iterations. For DOTA dataset, we trained 120k iterations, and the learning rate changed during the 40k and 80k iterations. Besides, weight decay and momentum are 0.0001 and 0.9, respectively. The optimizer chosen is MomentumOptimizer. Furthermore, we flip the image randomly in the training process, while subtracting the mean value $[103.939, 116.779, 123.68]$ which comes from ImageNet [22]. Subtracting mean can centralize all dimensions of the input data, which is conducive to model training. First of all, we use the SRSS to verify the feasibility of each part of the model, especially prow prediction, and then measure the overall performance and scalability of the model through the larger, more authoritative DOTA data set.

*B. Evaluation Indicators*

To quantitatively evaluate the performance of different framework in object detection, we use the precision–recall curve (PRC), mean Average Precision (mAP) and F-measure (F1), which are three well-known and widely applied standard measures approaches for comparisons [52].

PRC is obtained from four well-established evaluation components in information retrieval, true positive (TP), false positive (FP), false negative (FN), and true negative (TN) [46]. TP and FP indicate the number of correct predictions and error predictions. FN is the sum of regions not proposed. Based on these four components, we provide the definition of precision and recall rate as:

$$Precision = \frac{TP}{(TP + FP)} \tag{8}$$

$$Recall = \frac{TP}{(TP + FN)} \tag{9}$$

F1 is a statistic that is commonly used in the field of object detection. The higher the F1 value, the better the performance. The definition is as follows:

$$F1 = \frac{2 \times Precision \times Recall}{Precision + Recall} \tag{10}$$

mAP is to solve the single-point value limitations of P, R, F1, it can get an indicator that reflects global performance. The definition is as follows:

$$mAP = \int_0^1 P(R)dR \tag{11}$$

*C. Evaluation of DFPN*

As we all know, low-level feature semantic information is relatively scarce, but the object location is accurate. On the contrary, high-level feature semantic information is rich, but the object location is relatively rough. Therefore, the choice of feature maps is particularly important. In this section, we chose six different feature maps combination strategies to explore the impact on the detection performance. The specific combination strategies are shown in Table I. As we can see, the model achieved the worst detection performance when using only the P3 feature map. Furthermore, the combination of P3+P4 is significantly better than the combination of P2+P3 and the combination of P4+P5. This is due to the fact that most of the ships in our data match the anchors in P3 and P4. What's more, the P2 layer is mainly used for small object detection, P5 layer for large object detection. When using all feature maps, the detection performance is optimized: 85.2% for Recall, 84.5% for Precision, and 84.9% for F-measure.

TABLE I
PERFORMANCE OF DIFFERENT FEATURE MAPS COMBINATION STRATEGIES

| Combination Strategies | RECALL (%) | Precision (%) | F1 (%) |
|---|---|---|---|
| P3 | 72.7 | 80.1 | 76.2 |
| P2+P3 | 75.0 | 80.0 | 77.4 |
| P3+P4 | 81.6 | 82.1 | 81.8 |
| P4+P5 | 75.5 | 80.8 | 78.1 |
| P3+P4+P5 | 84.7 | 84.2 | 84.4 |
| P2+P3+P4+P5 | **85.2** | **84.5** | **84.9** |

TABLE II
PERFORMANCE OF THREE METHODS FOR EXTRACTING FIXED-LENGTH
FEATURE VECTORS

| Methods | RECALL (%) | Precision (%) | F1 (%) |
|---|---|---|---|
| ROI Align (R-DFPN) | 83.7 | 82.4 | 83.0 |
| RROI Align (RRPN) | 81.6 | 83.2 | 82.4 |
| Adaptive ROI Align | **85.2** | **84.5** | **84.9** |

TABLE III
DETECTION PERFORMANCE OF FIVE METHODS

| Methods | RECALL (%) | Precision (%) | F1 (%) | Time (s) |
|---|---|---|---|---|
| Faster-RCNN | 75.9 | 88.3 | 81.7 | 0.1 |
| FPN | 77.0 | **89.3** | 82.7 | 0.15 |
| FPN-Soft-NMS | 80.5 | 87.6 | 83.9 | 0.15 |
| LSTM-Based | 76.4 | 82.4 | 79.3 | **0.05** |
| Ours | **85.2** | 84.5 | **84.9** | 0.16 |

In summary, multiscale detection networks are significantly better than single-scale detection networks, especially in the detection of small objects. Only make full use of effective fusion of various layers of feature information, can we achieve better results.

### D. Evaluation of Adaptive ROI Align

Due to the large number of redundant regions in ship detection, the ultimate detection performance is often compromised. Fig. 6 shows three methods for extracting fixed-length feature vectors, while comparing their differences and their advantages and disadvantages. In this section, we will conduct specific experiments on these three methods, and parametrically compare the performance among them.

Table II shows the detection performance of ROI Align (R-

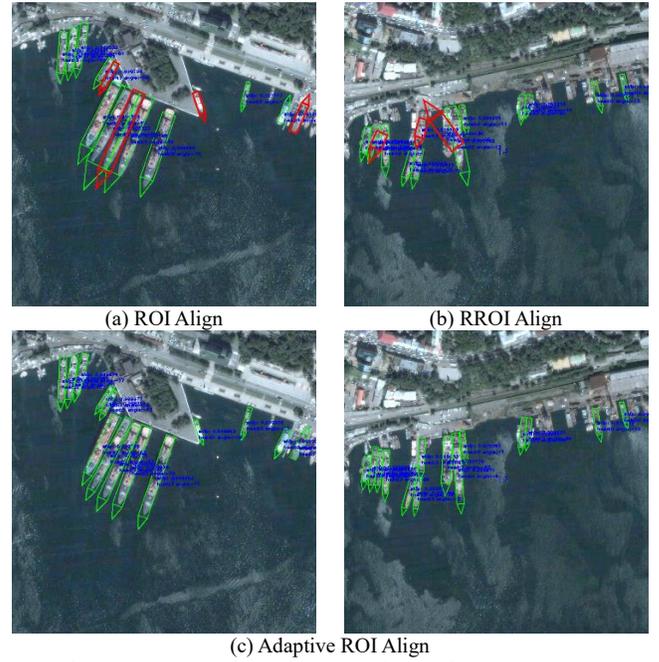

(a) ROI Align      (b) RROI Align

(c) Adaptive ROI Align

Fig. 7. Detection results of three Align methods. The red boxes and green boxes represent mission prediction boxes and correct detection boxes respectively. (a) ROI Align. (b) RROI Align. (c) Adaptive ROI Align

DFPN) [1], RROI Align (RRPN) [49] and Adaptive ROI Align. Obviously, adaptive ROI Align achieved the best results, especially the recall. Because the noise in the redundant regions often interfere with or even submerges the features, the ROI Align detection results often have missing detection and false alarms, as shown in Fig. 7(a). Although RROI Align completely eliminates the interference of redundant regions, it loses spatial information. The detection performance of RROI Align still unsatisfactory, which is reflected in the inaccurate prediction results and large angle deviation, as illustrated in Fig. 7(b). Because of the sensitive relationship between IoU overlap and rotation angle, the recall of RROI Align has been improved slightly. Fig. 7(c) shows adaptive ROI Align has the advantage of reducing the influence of noise and preserving the spatial information. Therefore, the detection result of adaptive ROI Align is accurate and the recall is high.

### E. Comparisons with Other Ship Detection Methods

In order to prove our proposed method is more competitive than traditional detection methods which are state-of-the-art in computer vision, we compare our proposed method with Faster-RCNN, FPN, FPN-Soft-NMS and LSTM-Based.

Table III show the quantitative comparison results of five methods, measured by F1. In the comparison of traditional detection methods (Faster-RCNN and FPN), FPN based on multiscale network has better performance. Meanwhile, FPN obtains the highest Precision value among (89.3%) the five methods. Soft-NMS [54] makes the bounding box, whose IoU exceeds the threshold have a certain probability to be reserved, so it is helpful to the dense scene detection. The results show that the FPN-Soft-NMS achieves an increase of about 1.2% without adding additional training and computational burden. LSTM-Based is a novel structure of the detection network, which introduces the LSTM structural unit. At the same time, it uses the Hungarian algorithm to serialize the output of the final detection result without any post-processing. This approach avoids the use of nonmaximum suppression operation and is suitable for use in dense scenes. However, this method still has certain limitations to highly overlapping objects, such as large aspect ratio ships. In the final detection results, detection performance of LSTM-Based is not prominent. Compared with the four detection framework above, the detection model proposed in this paper has achieved the best detection performance, and has the highest recall. Although our method offers superior performance in both multiscale and high-density object, we can see from Table III that the Precision of our method is not the highest, being behind that of the traditional method. This phenomenon shows that our method has a higher false alarm.

| Method | PL | BD | BR | GTF | SV | LV | SH | TC | BC | ST | SBF | RA | HA | SP | HC | mAP |
|--------|------|------|------|------|------|------|------|------|------|------|------|------|------|------|------|------|
| SSD | 41.06 | 24.31 | 4.55 | 17.1 | 15.93 | 7.72 | 13.21 | 39.96 | 12.05 | 46.88 | 9.09 | 30.82 | 1.36 | 3.5 | 0.0 | 17.84 |
| YOLOv2 | 52.75 | 24.24 | 10.6 | 35.5 | 14.36 | 2.41 | 7.37 | 51.79 | 43.98 | 31.35 | 22.3 | 36.68 | 14.61 | 22.55 | 11.89 | 25.49 |
| R-FCN | 39.57 | 46.13 | 3.03 | 38.46 | 9.10 | 3.66 | 7.45 | 41.97 | 50.43 | 66.98 | 40.34 | 51.28 | 11.14 | 35.59 | 17.45 | 30.84 |
| FR-H | 49.74 | 64.22 | 9.38 | 56.66 | 19.18 | 14.17 | 9.51 | 61.61 | 65.47 | 57.52 | 51.36 | 49.41 | 20.8 | 45.84 | 24.38 | 39.95 |
| FR-O | 79.42 | **77.13** | 17.7 | 64.05 | 35.3 | 38.02 | 37.16 | 89.41 | **69.64** | 59.28 | 50.3 | 52.91 | 47.89 | 47.40 | 46.30 | 54.13 |
| R-DFPN | 80.92 | 65.82 | 33.77 | 58.94 | 55.77 | **50.94** | 54.78 | 90.33 | 66.34 | 68.66 | 48.73 | 51.76 | 55.10 | 51.32 | 35.88 | 57.94 |
| Ours | **81.25** | 71.41 | **36.53** | **67.44** | **61.16** | 50.91 | **56.60** | **90.67** | 68.09 | **72.39** | **55.06** | **55.60** | **62.44** | **53.35** | **51.47** | **62.29** |

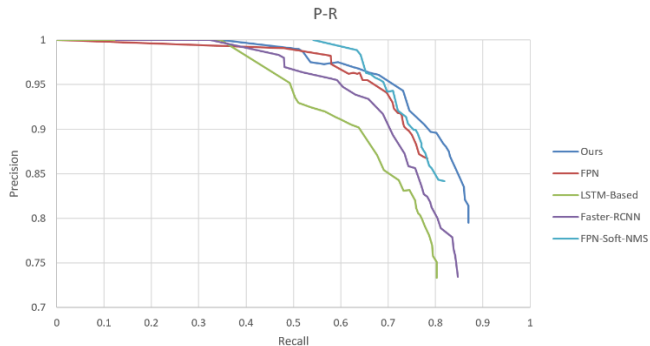

Fig. 8. The P-R curves of different methods. The proposed method has the state-of-the-art performance

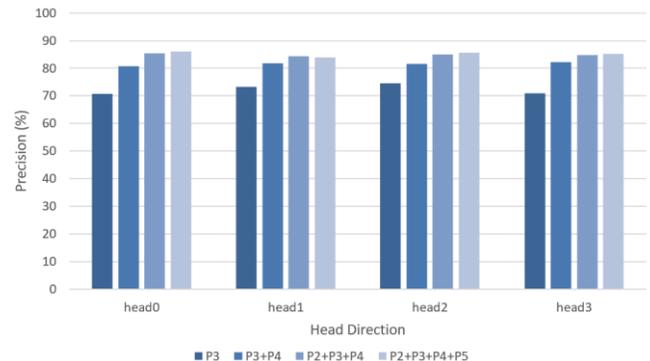

Fig. 9. Accuracy of prow direction prediction.

The time required for computing is summarized in Table III. LSTM-Based has the fastest detection speed. Although our method is the slowest of all detection algorithms, it is still very efficient.

Different Recall and Precision can be obtained by changing the confidence score threshold of detection. Fig. 8 plots the performance curves of different methods. As shown in the figure, the proposed method has the highest recall in a given precision. Similarly, with a specific recall, the proposed method has the highest precision. In short, the method we proposed has the best performance

### F. Evaluation of Prow Direction Prediction

In this section, we will estimate the accuracy of the prow direction prediction of our rotational region detection method.

Fig. 9 is the statistical results. The accuracy of prow predictions depends on the recall and we discover that the prediction accuracy in each direction is close to the recall, therefore it is a strong evidence that this simple prediction method is efficient and feasible.

An interesting phenomenon can be found in the experiment, that the final prediction often appears only in the prow and stern. We speculate that our network first learned that the prow must be in the long side and then judged which side of the long edge is the prow.

### G. Evaluation of overall performance and scalability

Apart from ship, our model can also detect multiple object categories. We evaluate the overall performance and scalability of our framework on a publicly available dataset, named DOTA, which is a large-scale dataset and it contains 15 common object categories. It should be noted that because the DOTA's annotation content is not suitable for the direction prediction of our model, some of the object categories have no concept of direction at all. Therefore, part of the direction prediction is blocked when we use the DOTA data set.

In Table IV, we shows the results obtained by using the horizontal region detection like SSD-inception-v2 [41], YOLOv2 [55], R-FCN [43], and FR-H [27], respectively. The results show that the mAP of ours are much higher than those who use others, especially in dense and large aspect ratio object detection such as ship, small-vehicle, large-vehicle, plane, harbor, bridge and so on. We also compare the results with FR-O [50] and R-DFPN [1], which are both based on rotation region detection. The detection accuracy of objects are improved to varying degrees through using our method and R-

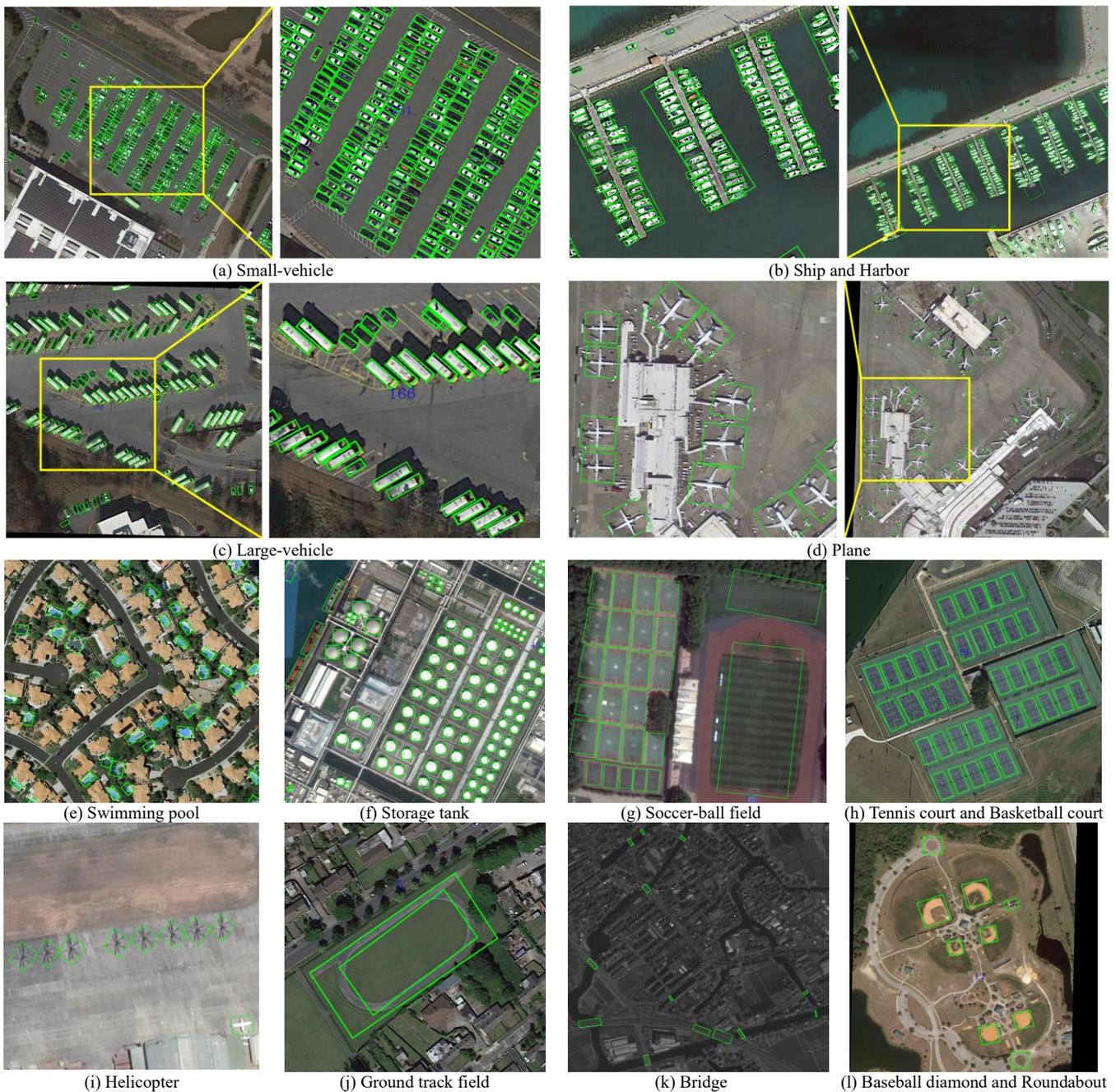

(a) Small-vehicle

(b) Ship and Harbor

(c) Large-vehicle

(d) Plane

(e) Swimming pool

(f) Storage tank

(g) Soccer-ball field

(h) Tennis court and Basketball court

(i) Helicopter

(j) Ground track field

(k) Bridge

(l) Baseball diamond and Roundabout

Fig. 10. Rotation region detection results on test data set of DOTA.

DFPN, which are primarily contributed to the use of DFPN. However, due to the use of adaptive ROI Align, our method has achieved better results than R-DFPN. As shown in Fig. 10, despite the large variations in the orientations and sizes of objects, the proposed approach has successfully detected and located most of the objects.

## V. CONCLUSION

In this paper, we build an end to end ship detection framework based on rotation regions which can handle different complex scenarios, detect intensive objects, and reduce redundant detection regions. Many novel structures were designed for this model. For example, we design a new multiscale feature fusion network, called DFPN, which can effectively integrate the low-level location information and high-level semantic information to provide more advanced features for object detection. Meanwhile, we explore the detection performance of different feature maps combination strategies. Then, adaptive ROI Align is proposed in this paper to mitigate the effects of redundant noise regions in the proposals and keep the completeness of semantic and spatial information. In addition, the berthing and sailing direction of ship has been found through prediction. At last, we adopt R-NMS which has more stringent constraints to obtain more accurate prediction. Experiments based on DOTA and SRSS dataset for rotation region detection show that our detection method has a competitive performance, as shown in Fig. 10 and Fig. 11.

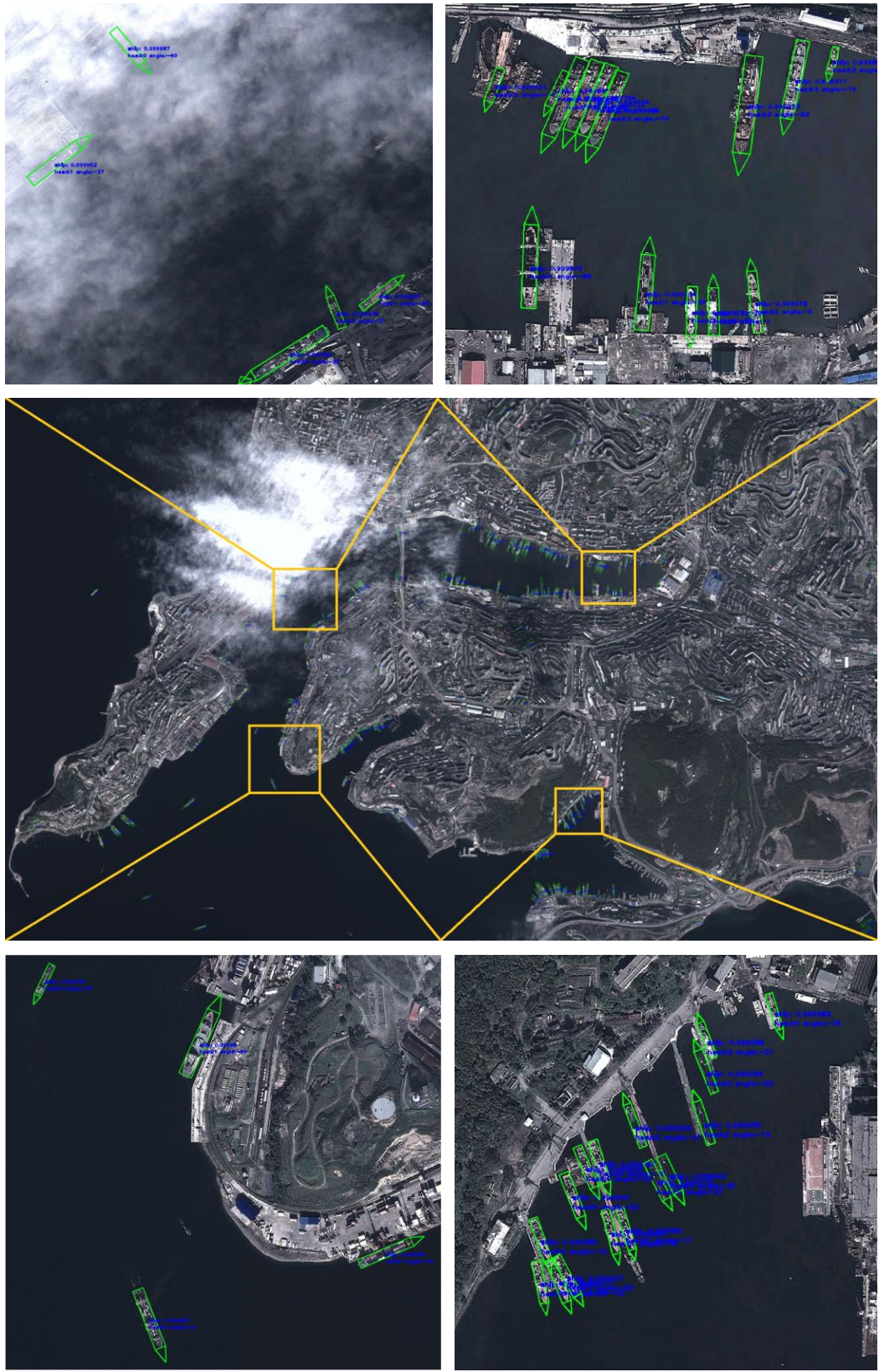

Fig. 10. The detection results of the proposed method near naval base.

Despite achieving the best performance, there are still some problems. More false alarms have resulted in a much lower

Precision than Faster-RCNN and FPN. We need to explore how to effectively reduce false alarms in the future.

ACKNOWLEDGMENT

The authors would like to thank all the colleagues in the lab, who generously provided their image dataset with the ground truth. The authors would also like to thank the anonymous reviewers for their very competent comments and suggestions.